\documentclass[conference]{IEEEtran}
\IEEEoverridecommandlockouts
\usepackage{cite}
\usepackage{amsmath,amssymb,amsfonts}
\usepackage{algorithmic}
\usepackage{graphicx}
\usepackage{textcomp}
\usepackage{xcolor}
\def\BibTeX{{\rm B\kern-.05em{\sc i\kern-.025em b}\kern-.08em
    T\kern-.1667em\lower.7ex\hbox{E}\kern-.125emX}}
\begin{document}

\title{\huge Computer Aided Diagnosis and Out-of-Distribution Detection in Glaucoma Screening Using Color Fundus Photography}

\author{\IEEEauthorblockN{Satoshi Kondo}
\IEEEauthorblockA{Muroran Institute of Technology \\
Hokkaido, Japan \\
kondo@mmm.muroran-it.ac.jp}
\and
\IEEEauthorblockN{Satoshi Kasai}
\IEEEauthorblockA{Niigata University of Healthcare and Welfare \\
Niigata, Japan \\
satoshi-kasai@nuhw.ac.jp}
\and
\IEEEauthorblockN{Kosuke Hirasawa}
\IEEEauthorblockA{Konica Minolta, Inc. \\
Osaka, Japan \\
kosuke.hirasawa@konicaminolta.com}
}

\maketitle

\begin{abstract}
Artificial Intelligence for RObust Glaucoma Screening (AIROGS) Challenge is held for developing solutions for glaucoma screening from color fundus photography that are robust to real-world scenarios. This report describes our method submitted to the AIROGS challenge. Our method employs convolutional neural networks to classify input images to "referable glaucoma" or "no referable glaucoma". In addition, we introduce an inference-time out-of-distribution (OOD) detection method to identify ungradable images. Our OOD detection is based on an energy-based method combined with activation rectification.

\end{abstract}


\section{Introduction}

Early detection of glaucoma can avoid visual impairment,  which could be facilitated through screening. Artificial intelligence (AI) could increase the cost-effectiveness of glaucoma screening, by reducing the need for manual labor. AI approaches for glaucoma detection from color fundus photography (CFP) have been proposed and promising at-the-lab performances have been reported. However, large performance drops often occur when AI solutions are applied in real-world settings. Unexpected out-of-distribution data and bad quality images are major causes for this performance drop.

Artificial Intelligence for RObust Glaucoma Screening (AIROGS) Challenge~\cite{airogs} is held to develop solutions for glaucoma screening from CFP that are robust to real-world scenarios. 

This report describes our method submitted to the AIROGS challenge. Our method employs convolutional neural networks to classify input images to "referable glaucoma" or "no referable glaucoma". In addition, we introduce an inference-time out-of-distribution (OOD) detection method to identify ungradable images. Our OOD detection is based on an energy-based method combined with activation rectification.

\section{Overview of AIROGS Challenge}

In this challenge, the Rotterdam EyePACS AIROGS dataset is used~\cite{eyepacs}. This dataset contains 113,893 color fundus images from 60,357 subjects and approximately 500 different sites with a heterogeneous ethnicity. All images is assigned by human experts with the labels of referable glaucoma, no referable glaucoma, or ungradable.

The training set provided by the organizers only consists of gradable images and ungradable images are excluded. In the test set, however, both gradable and ungradable images are included. Hence, the ungradable images cannot be used in the training phase.

For each input image, the following four outputs are expected; a likelihood score for referable glaucoma, a binary decision on referable glaucoma presence, a binary decision on whether an image is ungradable, and a scalar value that is positively correlated with the likelihood for ungradability.

\section{Proposed Method}

\subsection{Overview}

Our method employs convolutional neural networks to classify input images to "referable glaucoma" or "no referable glaucoma". In addition, we introduce an inference-time OOD detection method to identify ungradable images. Our OOD detection is based on an energy-based method combined with activation rectification. We will explain the details of the proposed method in the following sections.

\subsection{Classification}

We employ ResNet-RS~\cite{bello2021revisiting} for the classifying color fundus images to "referable glaucoma" or "no referable glaucoma". Since it is two class classification, a single linear layer with two channel output is attached at the end of the network instead of the final linear layer of ResNet-RS. 

\subsection{Out-of-distribution Detection}

Our inference-time OOD detection is based on an energy-based method~\cite{liu2020energy} combined with activation rectification~\cite{sun2021react}. The energy-based method uses a scoring function based on energy, instead of softmax, to discriminate in-distribution (ID) and OOD data. In the activation rectification, the outsized activation of a few layers can be attenuated by rectifying the activations at an upper limit. After rectification, the output distributions for ID and OOD data become much more well-separated. It is based on the observation that the mean activation for ID data is well-behaved with a near-constant mean and standard deviation, and the mean activation for OOD data has significantly larger variations across units and is biased towards having sharp positive values.

We consider a pre-trained neural network parameterized by $\theta$, which encodes an input $\mathrm{x}$ to a feature vector. We denote the feature vector from the penultimate layer of the network by $h(\mathrm{x}) \in \mathbb{R}^{m}$. The activation rectification operation is applied to the feature vector as:
\[
	\bar{h}(\mathrm{x}) = \min (h(\mathrm{x}), c),
\]
where the operation $\min$ is applied element-wise and $c$ is the activation threshold. The model output after the activation rectification is given by:
\[
	\bar{f}(\mathrm{x}; \theta) = \mathbf{W}^{T} \bar{h}(\mathrm{x}) + \mathbf{b},
\]
where $\mathbf{W}$ and $\mathbf{b}$ are the weight matrix and the bias vector of the last layer of the network. The energy of the model output is given by:
\[
	E(\mathrm{x}; \bar{f}) = -T \cdot \log \sum_{i}^{K} e^{\bar{f}_{i}(\mathrm{x}; \theta) / T},
\]
where $T$ is a temperature parameter and $\bar{f}_{i}(\mathrm{x}; \theta)$ indicates the logit corresponding to the $i$-th class label.

The OOD scoring function is given by:
\[
	g(\mathrm{x}; \tau, \bar{f}) =
	\begin{cases}
		\text{ID} & \text{if $-E(x; \bar{f}) \ge \tau$,} \\
		\text{OOD} & \text{if $-E(x; \bar{f}) < \tau$,} 
	\end{cases}
\]
where $\tau$ is the energy threshold. And we define the scalar value that is positively correlated with the likelihood for ungradability as $\tau + E(x; \bar{f})$.

\subsection{Model Ensemble}

We use multiple models as described in the next section. Each model is trained independently and the inference results are obtained by ensemble of the outputs from the models. The final likelihood score for referable glaucoma is obtained by averaging the likelihood scores (softmax) from the models. The binary decision on referable glaucoma presence is obtained by thresholding the final likelihood score. The final binary decision on whether the image is ungradable is obtained by majority voting of the models. And the final scalar value that is positively correlated with the likelihood for ungradability is obtained by averaging the scalar values from the models.

\subsection{Training Procedure}

We use 200-layer ResNet-RS pretrained with the ImageNet dataset~\cite{russakovsky2015imagenet} for our base network. 

As for the preprocessing, the images are cropped and resized to 256 $\times$ 256 pixels and then augmented. We use shift (maximum shift size is 10\,\% of the image size), scaling (0.9\,--\,1.1 times), rotation (-5\,--\,+5 degrees), color jitter (0.8\,--\,1.2 times for brightness, contrast, saturation and hue) and Gaussian blur (the max value of the sigma is 1.0) for the augmentation. The images are then normalized.

We use the cross entropy loss as the loss function. Since the dataset is imbalanced (there are much more "no referable glaucoma" samples than "referable glaucoma" samples), the weights are given depending on the inverse of the frequencies in the training dataset.

The whole dataset includes 101,442 images. We divide the dataset into five folds. Five models are trained by using different combinations of the folds. Four folds are used for training and the remaining one fold is used for validation. About 80,000 and 20,000 images are used in training and validation for each model, respectively.

The optimizer is Adam~\cite{dp2015adam} and the learning rate changes with cosine annealing. The initial learning rate is 1e-3. The number of epoch is 90. The model with the highest value in F1 score for the validation dataset is selected as the final model.

The thresholds $c$ and $\tau$ are decided by using the validation dataset for each model. For the activation threshold $c$, 90-th percentile of activations estimated on the validation dataset is used. For the energy threshold $\tau$, 95-th percentile of energies estimated on the validation dataset is used. We use the temperature parameter $T = 1$ .

\section{Experimental Results}

The organizers of the AIROGS challenge prepare the preliminary test site to evaluate the performance. The screening performance is evaluated using the partial area under the receiver operator characteristic curve (90\,--\,100\,\% specificity) for referable glaucoma and sensitivity at 95\,\% specificity.  The agreement between the reference and the prediction on image gradability is calculated by using Cohen's kappa score. Furthermore, the area under the receiver operator characteristic curve is determined using the human reference for ungradability as the true labels and the ungradability scalar values provided by the predictions as the target scores. 

The results of the preliminary test of our proposed method were that the screening sensitivity was 0.787, the screening partial AUC was 0.857, the ungradability kappa was 0.359, and the ungradability AUC was 0.863, 

\bibliographystyle{IEEEtran}
\bibliography{references}

\begin{thebibliography}{1}
\providecommand{\url}[1]{#1}
\csname url@samestyle\endcsname
\providecommand{\newblock}{\relax}
\providecommand{\bibinfo}[2]{#2}
\providecommand{\BIBentrySTDinterwordspacing}{\spaceskip=0pt\relax}
\providecommand{\BIBentryALTinterwordstretchfactor}{4}
\providecommand{\BIBentryALTinterwordspacing}{\spaceskip=\fontdimen2\font plus
\BIBentryALTinterwordstretchfactor\fontdimen3\font minus
  \fontdimen4\font\relax}
\providecommand{\BIBforeignlanguage}[2]{{%
\expandafter\ifx\csname l@#1\endcsname\relax
\typeout{** WARNING: IEEEtran.bst: No hyphenation pattern has been}%
\typeout{** loaded for the language `#1'. Using the pattern for}%
\typeout{** the default language instead.}%
\else
\language=\csname l@#1\endcsname
\fi
#2}}
\providecommand{\BIBdecl}{\relax}
\BIBdecl

\bibitem{airogs}
\BIBentryALTinterwordspacing
 [Online]. Available: \url{https://airogs.grand-challenge.org/}
\BIBentrySTDinterwordspacing

\bibitem{eyepacs}
\BIBentryALTinterwordspacing
``Rotterdam eyepacs airogs train set.'' [Online]. Available:
  \url{https://doi.org/10.5281/zenodo.5745363}
\BIBentrySTDinterwordspacing

\bibitem{bello2021revisiting}
I.~Bello, W.~Fedus, X.~Du, E.~D. Cubuk, A.~Srinivas, T.-Y. Lin, J.~Shlens, and
  B.~Zoph, ``Revisiting resnets: Improved training and scaling strategies,''
  \emph{Advances in Neural Information Processing Systems}, vol.~34, 2021.

\bibitem{liu2020energy}
W.~Liu, X.~Wang, J.~Owens, and Y.~Li, ``Energy-based out-of-distribution
  detection,'' \emph{Advances in Neural Information Processing Systems},
  vol.~33, pp. 21\,464--21\,475, 2020.

\bibitem{sun2021react}
Y.~Sun, C.~Guo, and Y.~Li, ``React: Out-of-distribution detection with
  rectified activations,'' \emph{Advances in Neural Information Processing
  Systems}, vol.~34, 2021.

\bibitem{russakovsky2015imagenet}
O.~Russakovsky, J.~Deng, H.~Su, J.~Krause, S.~Satheesh, S.~Ma, Z.~Huang,
  A.~Karpathy, A.~Khosla, M.~Bernstein \emph{et~al.}, ``Imagenet large scale
  visual recognition challenge,'' \emph{International journal of computer
  vision}, vol. 115, no.~3, pp. 211--252, 2015.

\bibitem{dp2015adam}
D.~P. Kingma and J.~Ba, ``Adam: A method for stochastic optimization,'' in
  \emph{Proc. of the 3rd International Conference for Learning Representations
  (ICLR)}, 2015.

\end{thebibliography}

\end{document}